\documentclass[conference]{IEEEtran}
\IEEEoverridecommandlockouts
\usepackage[plain,noend,ruled,linesnumbered]{algorithm2e}
\usepackage{amsmath,amssymb,amsfonts}
\usepackage{algorithmic}
\usepackage{graphicx}
\usepackage{subfig}
\usepackage{textcomp}
\usepackage{xcolor}
\usepackage{multirow}
\usepackage{mathrsfs}
\usepackage{physics}
\usepackage{mathtools}
\usepackage{booktabs} % for better formatting of tables
\usepackage{array}    % for better column formatting
\usepackage[numbers,sort&compress]{natbib}
\usepackage{color}
\def\BibTeX{{\rm B\kern-.05em{\sc i\kern-.025em b}\kern-.08em
    T\kern-.1667em\lower.7ex\hbox{E}\kern-.125emX}}

\begin{document}

\title{PFedDST: Personalized Federated Learning with Decentralized Selection Training}

\author{\IEEEauthorblockN{Mengchen Fan$^{a}$, Keren Li$^{b}$, Tianyun Zhang$^{c}$, Qing Tian$^{a}$ and Baocheng Geng$^{a}$}
\IEEEauthorblockA{$^a$\textit{Department of Computer Science, University of Alabama at Birmingham, Birmingham, US}\\
$^b$\textit{Department of Mathematics, University of Alabama at Birmingham, Birmingham, US}\\
$^c$\textit{Department of Computer Science, Cleveland State University, Cleveland, US}
}
}

\maketitle

\begin{abstract}

Distributed Learning (DL) enables the training of machine learning models across multiple devices, yet it faces challenges like non-IID data distributions and device capability disparities, which can impede training efficiency. Communication bottlenecks further complicate traditional Federated Learning (FL) setups. To mitigate these issues, we introduce the Personalized Federated Learning with
Decentralized Selection Training (PFedDST) framework. PFedDST enhances model training by allowing devices to strategically evaluate and select peers based on a comprehensive communication score. This score integrates loss, task similarity, and selection frequency, ensuring optimal peer connections. This selection strategy is tailored to increase local personalization and promote beneficial peer collaborations to strengthen the stability and efficiency of the training process. Our experiments demonstrate that PFedDST not only enhances model accuracy but also accelerates convergence. This approach outperforms state-of-the-art methods in handling data heterogeneity, delivering both faster and more effective training in diverse and decentralized systems.

\end{abstract}


\section{Introduction}
% {\color{blue} I changed the abstract, check if it is correct. we still need to add some references in the introduction.}
% Interest in leveraging signal processing and machine learning (ML) for critical predictive tasks across various fields is on the rise. The strategy of integrating information from multiple sources, such as sensors, for
% information gathering promises enhanced outcomes by offering multiple viewpoints on a single phenomenon. Additionally, the global shift towards stricter data privacy laws, with an increasing number of entities implementing regulations against the sharing of sensitive data like health information, has spurred progress in distributed learning and decision-making processes that operate without sharing raw data.
% Interest in leveraging signal processing and machine learning (ML) for critical predictive tasks across various fields is increasing. 
There has been growing interest in applying signal processing and machine learning (ML) to critical predictive tasks across various disciplines \cite{zhang2019fusion,10706324,li2023nn,zhai2025machine,zhao2024deep,zhai2024impact}. The strategy of integrating data from multiple sources, such as sensors, enhances outcomes by providing multiple perspectives on a single phenomenon. Moreover, the global trend toward stricter data privacy laws and the growing implementation of regulations that restrict the sharing of sensitive data, such as health information, has accelerated advancements in distributed learning and decision-making processes that function without exchanging raw data \cite{quan2023distributed,quan2023efficient,quan2023ordered,geng2021collaborative,quan2022enhanced}.

Federated learning (FL) has gained prominence for its ability to train models on decentralized devices \cite{chen2021distributed}. FL systems facilitate multi-client learning without centralizing raw data, addressing both privacy and communication challenges. However, \textit{heterogeneity} in data distribution, resource allocation, task objectives, or network characteristics across nodes poses challenges to model accuracy and convergence \cite{fan2025measuringheterogeneitymachinelearning}. Personalized Federated Learning (PFL) \cite{tan2022towards} addresses these issues by tailoring models to specific client needs, thereby enhancing the effectiveness beyond the conventional single-model approach (e.g., \cite{arivazhagan2019federated}) in FL.

PFL is categorized into Centralized Personalized Federated Learning (CPFL) and Decentralized Personalized Federated Learning (DPFL) \cite{lalitha2018fully}. CPFL can suffer from communication bottlenecks and server failures, leading to increased communication traffic and potential system crashes. In contrast, DPFL emphasizes peer-to-peer interactions among edge clients, reducing communication loads on local nodes and promoting faster convergence. In this topology, clients maintain an undirected and symmetric communication structure, facilitating model exchanges with peers.
% In order to satisfy the unique needs of individual clients,
% most existing works in PFL carefully designed the relationships between the global model and personalized models to
% fit the local data distribution via different techniques, such
% as parameter decoupling , knowledge distillation, multi-task learning , model interpolation  and clustering. These techniques can
% also be adopted to improve the personalized performance in
% DPFL. However, the heterogeneity among clients
% exists not only in local data distribution but also in the communication power and computation resources. The
% power level of the wireless channel among clients may be
% different and time-varying in communication networks, and
% some clients may get offline occasionally without sending
% messages to their neighbors. These result in long-term waits
% or incidents of deadlock for their neighbors and also
% lead to poor convergence for the whole system. Besides,
% there is no reason to expect that the exchanged models are
% trained at the same convergence level due to the heterogeneous computation resources. Clients may receive excessive poor-performing models which can not help their training and degrade the personalized performance.

Most existing PFL approaches finely tune the interactions between global and personalized models to accommodate local data variations using methods such as regularization \cite{li2020federated}, knowledge distillation \cite{gou2021knowledge}, multi-task learning \cite{marfoq2021federated}, and clustering \cite{sattler2020clustered}. These techniques aim to enhance personalized performance in the heterogeneous setting. For example, approaches like FedPer \cite{arivazhagan2019federated} propose to capture personalization aspects in
federated learning by viewing deep learning models as base and personalization layers. And FedBABU \cite{oh2021fedbabu} utilize a single global feature representation coupled with multiple local classifiers, differing in how they manage the relationship between the shared representation and the individual linear components. FedFusion \cite{10706324} utilizes a representation method to fuse the batch information to solve the heterogeneity problem. Cho et al. \cite{cho2020client} provides theoretical convergence analysis for these algorithms under general non-convex conditions. DFedAvgM \cite{sun2022decentralized} employs multiple local iterations with SGD and quantization techniques to reduce communication overhead. Dis-PFL \cite{dai2022dispfl} designs personalized models and pruned masks for each client to personalized convergence. OSGP \cite{assran2019stochastic}, DfedPGP \cite{liu2024decentralized}, and AsyNG \cite{chen2023enhancing} utilize the push-sum method to enhance training efficiency.

Despite ongoing efforts, DPFL methodologies continue to face slow convergence rates during aggregations, a challenge compounded by heterogeneous data distributions among clients. Additionally, \textit{disparities in communication bandwidth and computational capabilities} complicate these issues further, leading to unstable communication channels between clients. As a result, clients are compelled to selectively engage with only a limited subset of peers for communication.%Furthermore, the power levels in client communication channels can fluctuate over time, complicating network stability. %Additionally, clients may occasionally go offline unexpectedly, failing to send updates to their neighbors. These issues can lead to prolonged waiting times or even deadlocks, affecting the convergence of the entire system. Moreover, due to varying computational resources, there is no guarantee that models exchanged between clients are trained to the same level of convergence. Consequently, clients may receive models that do not contribute positively to their training efforts and thus affect personalized performance.
% To tackle the data heterogeneity and client heterogeneity
% problems together, we propose PFedDST, a Decentralized
% selection training based Personalized Federated Learning
% approach, to solve the personalized FL problem using score strategy to find the benefit communication choices in the decentralized setting.
% Instead of deploying the consensus model for each user,
% PFedDST allows each client to own their personalized unique
% model's header, allowing them
% to better adapt to their local data. Specifically, the decentralized selection training technique mainly consists of three steps: First, the active client will grab the information in the client pool and use score strategy, which contains the information with the header and loss, and then combine the selection history to got a score for each avaliable client. Second, after the each client have final dicision, the local client need to directed collect the whole of feature extraction information from selected clients. Third, do the feature extraction training and push to the out neighbors and then do the header training and push to the selection pool and update the information. 

To address these challenges, we introduce Personalized Federated Learning with
Decentralized Selection Training (PFedDST), a decentralized selection training-based Personalized Federated Learning approach. This method ensures that each client maintains a model of the same dimensionality, facilitating efficient aggregation and strategic communication among clients. During each communication round, clients selectively engage with a subset of peers, chosen through a \textit{strategic scoring strategy} for their relevance to the current learning context. They then aggregate their own model with those selected from their peers. After local updates, clients share their newly trained model parameters with the required peers, thereby enhancing the collective learning process and ensuring continuous improvement and relevance of the shared data.
We employ an \textbf{innovative} scoring scheme that evaluates potential peer clients based on three key factors: feature extraction capability, task heterogeneity, and communication frequency. Simulations in heterogeneous settings demonstrate that PFedDST not only increases the average test accuracy on local test data but also reduces the number of communication rounds required to achieve the same performance targets.

%This method utilizes a scoring strategy to identify beneficial communication choices in a decentralized setting, aiming to solve personalized FL issues more effectively. Unlike conventional approaches that deploy a consensus model for each user, PFedDST enables each client to maintain a personalized model header, enhancing their ability to adapt to local tasks. This approach allows the active client first to collect and analyze information from the client pool, including frequency, headers, and losses. It employs a scoring strategy that integrates this information to assign scores to available clients. Once the local client makes a decision based on these scores, it directly gathers all necessary feature extraction information from the chosen peers. The client then proceeds with feature extraction training and push to the out-neighbors. Next, the client local updates its model header based on the newly trained parameters and sends this updated information to the selection pool, ensuring continuous improvement and relevance of the data shared among clients. The overall framework is shown in Figure (\ref{fig:overall}). In heterogeneity settings, the results demonstrate that PFedDSTincreases the averaged test accuracy on local test data, requires fewer communication rounds to reach the same target. 
We summarize our contributions as following: 
\begin{itemize} \item We propose the PFedDST framework, a personalized federated learning approach where each client continuously learns from selected peers to update its feature extraction capabilities while maintaining a personalized prediction header. This integration of peer selection and partial model personalization enhances robust communication and accelerates convergence. \item Strategic selection enables clients to enhance their feature extraction capabilities from the most informative and relevant neighbors. It also prioritizes communication with clients that have not recently interacted, thereby diversifying and refreshing the learning inputs. \item Experimental results demonstrate that PFedDST outperforms various state-of-the-art baselines. It proves particularly effective in environments characterized by data heterogeneity and limited computational resources. \end{itemize}

It should be noted that our strategy is different from traditional directed DFL methods such as Dis-PFL and AsyNG, which typically involve exchanging all parameters for a single consensus model or selecting communication targets randomly. Instead, our approach incorporates score-based neighbor selection, partial freeze \cite{brock2017freezeout} training, and alternating optimization to accelerate convergence. This method not only ensures model robustness and enhances personalization but also optimizes communication efficiency.

\section{System Model and Methodology}

In centralized model training, consider a classification or multiclass detection task in which each data sample is a pair of \((x,y)\), where $x \in \mathbb{R}^d$ represents the input features, and $y \in \{0, 1, \dots, k-1\}$ signifies the corresponding labels, with $k$ being the number of possible classes. The goal is to classify the variable \(x\) into one of \(k\) categories. This classification is achieved using a model parameterized by $w: \mathbb{R}^d \rightarrow \mathbb{R}^k$. Each component of the output, \(\gamma_y(x)\) for \(y = 0, \dots, k-1\), represents the likelihood (or confidence score) that the instance \(x\) belongs to class \(y\). The primary objective is to minimize the expected loss, defined by the equation:
\begin{equation}\label{eq:classical}
    \mathcal{L}(w) := \mathbb{E}_{(x,y) \sim D}[L(w;x, y)],
\end{equation}
where  \(L(w; x,y)\) measures the loss of the decision margins \(\gamma_y(x; w) \in \mathbb{R}^k\) when the true label of \(x\) is \(y\), and the expectation is taken over the joint distribution of the dataset $D$.

Optimization of this expected loss commonly uses gradient-based algorithms like Stochastic Gradient Descent (SGD) or Adam. These methods iteratively adjust the parameters $w$ to minimize the empirical loss for data $(x,y)$, $w_{\text{new}} = w_{\text{old}} - \eta \nabla_w L(w; x, y)$,
where $\eta$ denotes the learning rate.

\subsection{Decentralized Personalized Federated Learning with Partial Freezing}

In decentralized personalized federated learning, where data distribution varies across clients, each client $i$ has a distinct data distribution $D_i$ and maintains a personalized model parameterized by $w_i$. The index $i$ ranges over a total of $M$ clients, $i \in \{1, \dots, M\}$. In this setting, a objective is to optimize the local models jointly \cite{hanzely2021personalized}:
\begin{equation}\label{eq:obj}
\min_{w_1, \cdots, w_M} \mathcal{L}(w_1, \cdots, w_M) = \frac{1}{M}\sum_{i=1}^M \mathcal{L}_i(w_i),
\end{equation}
where $\mathcal{L}_i(w_i) = \mathbb{E}_{(x,y) \sim D_i} \left[ L(w_i; x, y) \right]$ represents the empirical risk associated with the $i$-th client's local data and $L$ denotes the loss function.

% To improve personalized performance, we consider the selection of partial-freeze in decentralized personalized FL. The model structure can consist of a fully personalized header $h_i$ and an aggregated feature extraction $e_i$, aggregating from the optimized selected neighbors $\mathcal{M}_i$. In detail, we divide the model into two parts: the header and the feature extraction layers. The header refers to the model's final linear layers, which primarily perform the final linear classification of the output. The feature extraction component comprises the remaining layers of the model, which are principally responsible for extracting features from the input data. During local optimization, these two sections are trained separately. Once the training of the feature extraction component is complete, it can then proceed to the next round of communication with out-neighbors. 

To enhance personalized model performance and expedite convergence, we integrate the concept of partially frozen training. \textit{The model is structured into two distinct parts: the header and the feature extraction layers.} The header comprises the model’s final fully-connected layers, primarily responsible for classification tasks. This component is essential for customizing the model to meet each client's specific requirements, allowing personalized adjustments in the decision-making process. The feature extraction layers consist of the earlier stages of the model, which are tasked with processing and extracting pertinent features from the input data.

During each communication round, each client $i$ strategically selects a subset of peer clients, denoted by $\mathcal{M}_i$, and aggregates (e.g., simple average) its own feature extraction layers with those from its peers to obtain the aggregated feature extraction layer $e_i$. The header layers of client $i$, denoted by $h_i$, remain unchanged and do not participate in the model aggregation. Then, client $i$ undergoes local training to sequentially update $e_i$ (with frozen $h_i$) and $h_i$ (with frozen $e_i$). Specifically, $h_i$ is frozen first and the aggregated $e_i$ is updated using the local data distribution $D_i$.

\begin{equation}\label{eq:obj_e}
\min_{e_i}\ \ \ \  \mathcal{L}_i(e_i) = \mathbb{E}_{(x,y) \sim D_i} \left[ L((e_i, h_i^f); x, y) \right]
\end{equation}
where the subscript $f$ in $h_i^f$ indicates that the parameters are frozen.
Upon updating $e_i$, it is sent to the required peers and the parameters in $h_i$ are unfrozen and updated next:

\begin{equation}\label{eq:obj_h}
\min_{h_i}\ \ \ \  \mathcal{L}_i(h_i) = \mathbb{E}_{(x,y) \sim D_i} \left[ L((e_i^f, h_i); x, y) \right]
\end{equation}

Once the local update is complete, client $i$ shares its updated $h_i$ back to the network. This updated information helps other clients make informed decisions about which peers to communicate with in the next round. The entire workflow of our approach is depicted in Figure 1.

\begin{figure}[ht]
\centering
\includegraphics[width=0.4\textwidth]{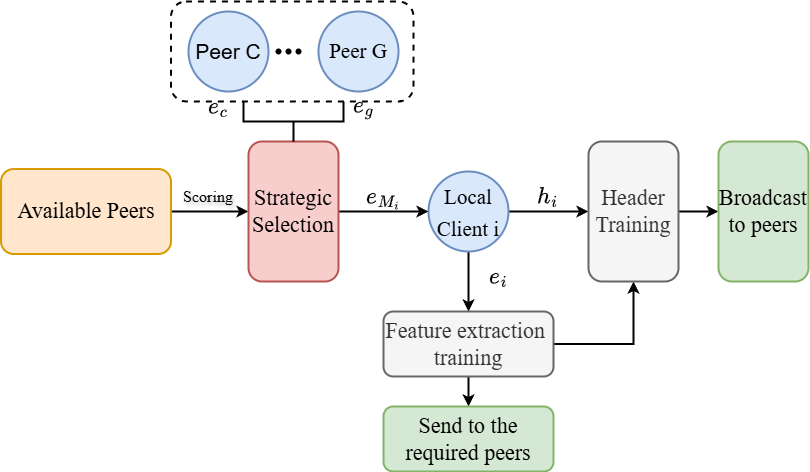} % Replace 'example-image.jpg' with your file's name
\caption{Overview of PFedDST} % This sets the image's title/caption
\label{fig:overall} % This labels the image for cross-referencing
\end{figure}

\subsection{Strategic Peer Selection for Communication}
For each client, we quantify the degree of information contribution from others by assigning a score to each peer. A higher score indicates that a peer holds more valuable information for enhancing the feature extraction capabilities of the local client, thus facilitating more targeted updates. Essentially, our aim is for clients to augment their feature extraction abilities from peers that are better equipped to guide them, particularly those undertaking similar tasks.

The scoring system is based on a composite evaluation of three factors: the loss disparity score $s_l$, the header distance $s_d$, and the peer recency $s_p$. Specifically:
\begin{itemize}
    \item The \textit{loss disparity score} ($s_l$) measures the potential to enhance the generality of a client’s feature extraction capabilities. This score is calculated by assessing the loss when predictions are made using this client’s model on a peer's local dataset. A higher loss indicates a significant gap in the client’s ability to predict the peer’s data, signaling a stronger need for adaptation.
    \item The \textit{header distance score} ($s_d$) identifies peers whose tasks are more closely related to the current client's tasks. It is measured by the weight distance between the header layers of the two clients. A smaller distance indicates that the label distributions, or tasks, of the two clients are more similar, making learning from such peers particularly beneficial.
    \item The \textit{peer recency score} ($s_p$) is designed to enhance learning generalization by avoiding repetitive communication with the same few peers and encouraging engagement with those not recently communicated with. This approach helps prevent overfitting and promotes a more diverse and robust learning process.
\end{itemize}

By employing this holistic scoring mechanism, we strategically select the most beneficial peers for communication, thereby optimizing the efficiency and effectiveness of the distributed learning environment.

\textbf{Loss Disparity Score.} 
The concept of selection skew, denoted by $\rho$, was defined in \cite{cho2020client} within the context of centralized FL. This skew quantifies the disparity in loss outcomes when evaluating the unified model on the data of a strategically selected subset of clients, as opposed to a random selection. The findings in \cite{cho2020client} suggest that a larger lower bound on $\rho$ leads to faster convergence during the training process, indicating the advantage of selecting clients whose data produce larger losses because the current model underperforms on these and requires further training.

Inspired by this, we define a \textit{decentralized version of $\rho$} for a specific local client $i$, representing the model loss difference between selecting a subset of peers $\mathcal{M}_i$ and a full random selection of peers in model aggregation:
\begin{equation}
    \rho_{i} = \frac{\sum_{j \in \mathcal{M}_i} n_j \left(\mathcal{L}_j(w_i) - \mathcal{L}_j(w_j^*)\right)/\sum_{j \in \mathcal{M}_i} n_j}{\mathcal{L}_i(w_i) - \sum_{j \in \mathcal{M}} n_j\mathcal{L}_j(w_j^*)/\sum_{j \in \mathcal{M}} n_j} \geq 0 
\label{eq:rho}
\end{equation}
where $\mathcal{M}$ represents the collection of available clients to the client $i$, $n_j$ is the fraction of data at the j-th client and $w_j^* = \arg \min_{w_j} \mathcal{L}_j$ is the optimized $w_j$ for client $j$. With purely random selection, $\rho = 1$ since the numerator and denominator in (\ref{eq:rho}) are equal.

Inspired by the findings in \cite{cho2020client}, we adopt a client selection strategy that seeks to maximize lower bound of $\rho$, thereby accelerating the convergence rate. However, evaluating $\rho$ for all potential subsets of peers is computationally impractical due to its NP-hard nature. Instead, we utilize the loss of applying the $i$th client's model on the $j$th peer's data, denoted by $l_j(w_{i,j})$, as a surrogate to measure the desirability of selecting peer $j$. Mathematically, the loss score between client $i$ and its peer $j$ is given by:
\begin{equation}
    s_l^{i,j} = \norm{l_j(w_{i,j})} = \norm{\mathbb{E}_{(x,y) \sim D_j} [ L(w_i;x,y) ]}
    \label{eq:sl}
\end{equation}
where $D_j$ is the data distribution at the $j$th peer. A higher $s_l^{i,j}$ indicates that the $i$th client's model struggles with handling the $j$th peer's data, suggesting a greater preference for selecting $j$ in the next communication round.

\textbf{Header Distance Score.} 
% \textit{Second, we also need to consider the influence of the data heterogeneity.} 
% Unlike the traditional centralized FL, which wants to obtain a highly generalized model to adapt the different type of data, the decentralized personalized FL mainly focus on obtain a highly personalized model to adapt to their local data. When we using the unbiased neighbor selection strategy, \textit{we noticed that} there are multiple selected neighbors have very low validation accuracy in the training process, show in Fig \ref{fig:validation}(a). The result show that most of selected neighbors have the low accuracy under the local validation data. Motivated by this result, we prefer selecting the neighbors have similar data distribution. Instead of validating the neighbors' parameters, which is very time-consuming, we want to use the distance of fully personalized headers to represent the data distance. 
Unlike traditional centralized FL, which aims to develop a unified model across diverse data types, decentralized personalized FL focuses on creating models tailored to specific local data. When two clients have similar tasks (e.g., comparable label distributions), they are likely to benefit from communication and model aggregation. Similar tasks imply compatible data or learning objectives, enhancing the learning process through effective information sharing. However, if two clients have significantly different tasks and data distributions, such as one client focusing on images of animals and another on images of plants, their feature extraction layers possess distinct properties. Aggregating their models might not only fail to improve but could potentially deteriorate each other's performance.

Based on this reasoning, it is preferable for clients to select peers with similar tasks. We propose using the element-wise cosine similarity between header layers' weights to measure this similarity. We prefer cosine similarity over Euclidean distance as a metric because it emphasizes the directional trends (patterns) of the weights rather than their absolute magnitudes. This approach is preferred as it focuses on the relative importance of input features, which reflects the true nature of the task. In addition, it is important to note that other distance metrics like Kullback-Leibler (KL) divergence are unsuitable for measuring distances between model parameters, as these weight parameters do not inherently possess probabilistic properties.

Let $H = (h_1, h_2, \dots, h_n)$ and $G = (g_1, g_2, \dots, g_n)$ represent the weight parameters corresponding to the header layers and the header distance score (coefficient) can be computed as follows:
\begin{equation}
  s_d = \frac{\sum_{i=1}^{n}h_i \cdot g_i}{\sqrt{\sum_{i=1}^{n}h_i^2} \cdot \sqrt{\sum_{i=1}^{n}g_i^2}} 
  \label{eq:sd}
\end{equation}
where $h_i$ and $g_i$ are the $i$-th elements of $H$ and $G$, and $n$ is the number of elements in $H$ and $G$.

%Unlike traditional centralized FL, which aims to develop a generalized model adaptable to various types of data, decentralized personalized FL focuses on creating personalized models tailored to local data. Communication between clients with the related task can enhance the efficiency and accuracy of local personalized training. Clients that have similar tasks often have compatible data or learning objectives, which can improve each other's learning processes through effective information sharing. Communication between clients with different tasks might be less beneficial or even detrimental to the training process. Based on this, selecting peers with similar tasks is more efficient. Instead of conducting time-consuming validations of peer parameters, we propose using the euclidean distance between fully personalized headers as a measurement for header distance. This method provides a more efficient way to validate the similarity of tasks between different clients, thereby enhancing the personalization and effectiveness of the model for each specific client.

\begin{figure}[ht]
    \centering
    \subfloat[Random selection]{ \includegraphics[width=0.4\textwidth]{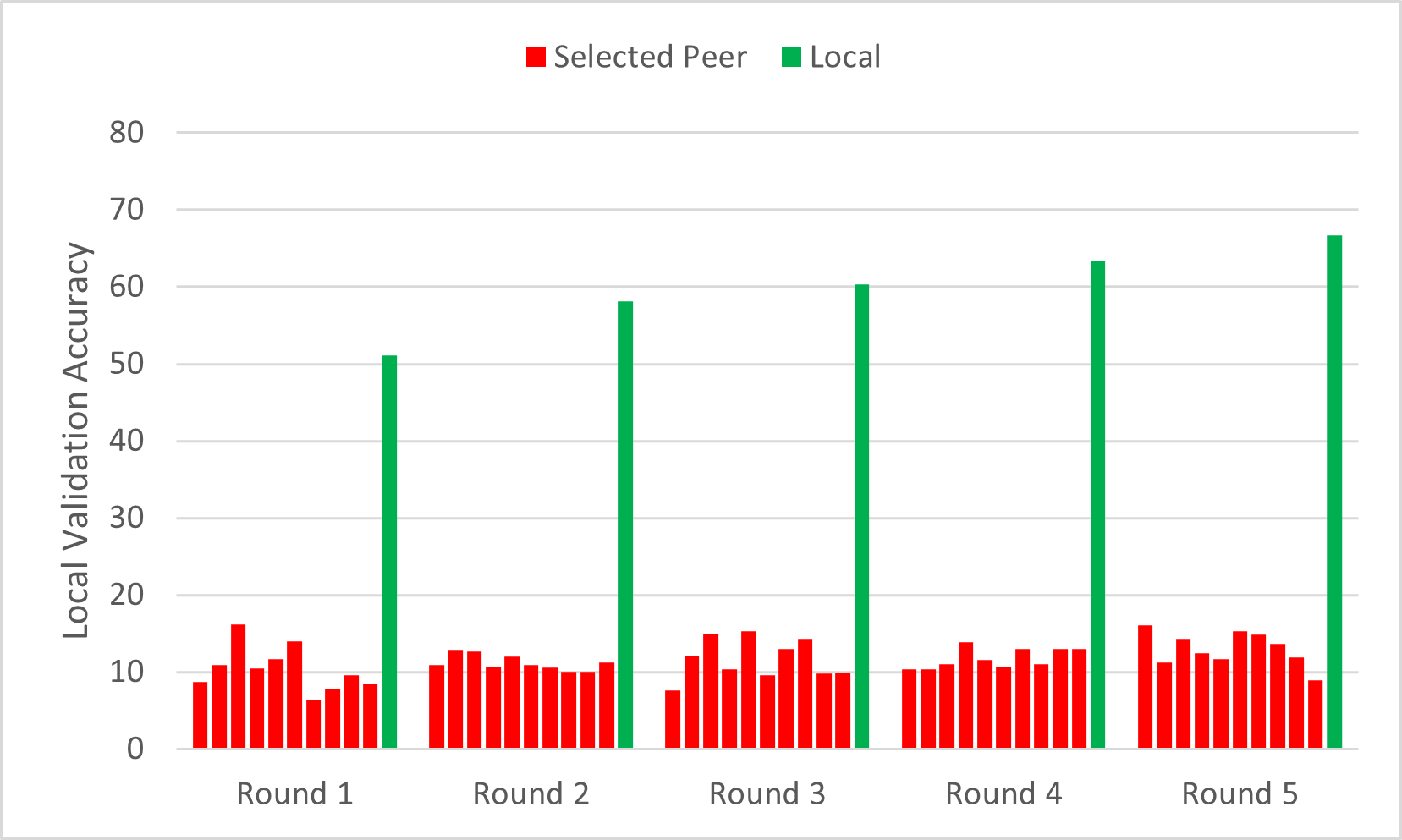}}

    \vspace{0.1cm}

   \subfloat[Score-based selection]{   \includegraphics[width=0.4\textwidth]{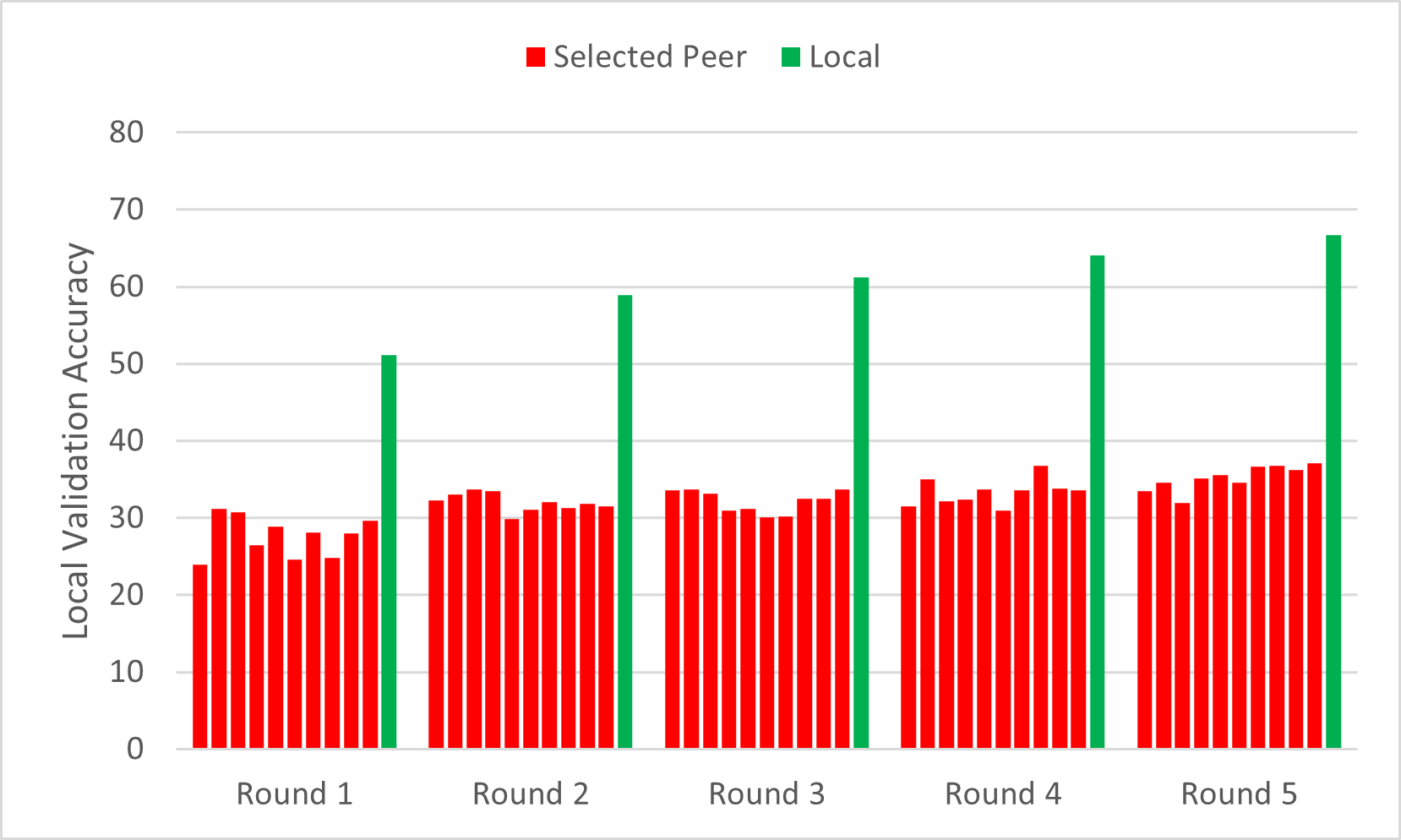}}

    \caption{The validation result of each selected peer in the local data}
    \label{fig:validation}
\end{figure}

Here, we validate the effectiveness of client selection based \textit{solely} on the header distance score. Figure 2 illustrates the model accuracy for a specific local client during decentralized training (for more details on the experimental setup, see the Experiments Section III). Each training round involves the client selecting 10 peers as candidates for communication. We evaluate the performance of the models from 10 selected peers on the local client's data, comparing random and strategic selections. Random selection is depicted in Figure \ref{fig:validation}(a), while strategic selection based on the header distance score is illustrated in Figure \ref{fig:validation}(b). The prediction accuracy of the models from the selected peers (represented by red bars) on the client’s data is plotted on the y-axis for each training round across both selection approaches.
While the local model performs best on its own data (denoted by green), models from strategically selected peers generally outperform those from peers selected randomly. This comparison demonstrates the utility of the header distance score in identifying the most relevant peers for communication.

%During the training process, we use local data to validate the effectiveness of each selected peer, represented by their accuracy. Figure \ref{fig:validation}(a) shows that neighbors selected through random selection generally have lower accuracy during the training process, which significantly reduces training efficiency. In contrast, after implementing our score-based selection method, each peer makes a useful contribution to the prediction of local data, demonstrating the utility of our approach in improving the training process, as shown in Figure %\ref{fig:validation}(b).

% {\color{red}give high level explaination of the meaning of this score}
\textbf{Peer Recency Score.} A critical factor in decentralized FL is the communication frequency. The peer recency score helps determine a local client's priority for selecting a peer based on how recently local client has communicated. This score is designed to prevent the local client from ``forgetting" the knowledge it might have acquired from peers that have not been engaged for several rounds. A higher peer recency score, indicating a longer interval since last communication, increases a peer's probability of being selected. This mechanism aims to enhance both the model's convergence rate and its generalization, thereby improving overall training effectiveness.

In a fully decentralized network, global training information and iteration counts are not accessible for each peer. Therefore, we calculate the peer recency score using only local iteration data. For a local client, let $n_{0,j}$ be the iteration number at which peer $j$ was last selected, and $n_t$ denote the current iteration number. The peer recency score of peer $j$, $s_{t,j}$ is designed to range from 0 to 1, where it approaches 0 if \( n_t - n_{0,j} \) is small, discouraging the repetitive selection of the same peer and promoting diversity in peer engagement. Conversely, as \( n_t - n_{0,j} \) exceeds a certain threshold \( c_0 \), $s_{t,j}$ increases to its maximum value of 1. To achieve this property, we use the cumulative distribution function (CDF) of the exponential distribution:

\begin{equation}
\label{eq:st}
    s_p = \phi(1 - e^{-\lambda (n_t - n_{0,j})})
\end{equation}
where $\phi$ represents the CDF and $\lambda$ is the rate (scaling) parameter of the exponential distribution.

%Since we focus on the fully decentralized network, the global training information and iteration is unavailable for each clients. Therefore, we propose this feasible solution to calculate the frequency score, which can be easily collected through the local client and can directly measure the peers priority. By adjusting the rate parameter $\lambda$ to be larger, we can make the $s_p \approx 1$ when the $n_i-n_0>c_0$, where $c_0$ is a threshold. %After $c_0$ times the neighbors are not be selected, we want the frequency score will not affect its selection priority.

\textbf{Holistic Determination of the Final Score.} For a specific client, the cumulative score for selecting a peer $j$ indicating its preference for selecting it incorporates three factors: the loss score $s_l$ from (\ref{eq:sl}), header distance score $s_d$ from (\ref{eq:sd}), and peer recency score $s_p$ from (\ref{eq:st}). For a specific local client, a peer's overall communication score is intelligently designed as follows,
\begin{equation}
\label{eq:score}
    \mathcal{S} = s_p(\alpha s_l - s_d + c)
\end{equation}
where $\alpha$ is a scaling parameter, and $c$ is a constant that represents the communication cost score between the corresponding peer.

This overall score increases under the following conditions: a. when $s_l$ increases, indicating a larger loss on the peer’s data and a greater need for the client to learn from it; b. when $s_d$ decreases, reflecting a higher task similarity with the peer; and c.  when $s_p$ increases, suggesting that the client has not communicated with this peer recently. In addition, the peer recency score $s_p$, ranging from 0 to 1, converges quickly to 1 as $|n_t - n_{0,j}|$ grows large. The multiplication of $s_p$ and $\alpha s_l - s_d + c$ ensures that $s_p$ does not dominate the selection process. This design prevents the selection of peers that are significantly different from the local client solely based on infrequent prior communication, therefore we enhance the stability of personalized training.

%We can know that \( s_p \) is a number between 0 and 1. When \( s_p \) approaches 0, which occurs if $n_i-n_0$ is small, \( s_p \) significantly influences the final score. For example, if a peer node was recently selected in the previous round, it will have a very small \( s_p \), thereby reducing its overall score substantially. This dynamic discourages the repeated selection of the same node, promoting diversity in node participation.

%Conversely, when \( n_i - n_0 \) exceeds \( c_0 \), the terms \( s_d \) and \( s_l \) become the dominant factors in the score, facilitating faster convergence. For instance, if a peer has not been selected for several iterations, \( s_p \) will approach 1, making the score dominant by \( s_d + \alpha s_l \). This approach rewards nodes that have been less active recently, providing them a higher score and thus a greater possibility of being selected in the current round.

%This mechanism allows us to find a trade-off between the speed of convergence and the error rate, thereby balancing these two critical aspects effectively. By adjusting the influence of \( s_p \), \( s_d \), and \( s_l \) based on the conditions of selection frequency and data distribution, we can optimize the overall performance of the distributed learning network. This strategic balancing act ensures that each relative node has an equitable chance of contributing to the model’s learning process, enhancing the robustness and personalization of the model.

\subsection{Algorithm}

In this section, we propose the PFedDST algorithm, which facilitates peer selection under a fully decentralized setting. The algorithm is designed to operate on each client, allowing for local decision-making without centralized oversight.

Each client maintains two context information arrays to support decision-making processes, the loss array \(l\) and the peer recency array \(t\).  The loss array \(l\) stores the loss information calculated from aggregated parameters with each peer, and the peer recency array \(t\) records the iteration numbers that each peer was last selected by the local client. By leveraging data from loss, header distance, and selection frequency, each eligible client is assigned a score that reflects its priority as a potential peer. This score is then used to determine the selected communication peers $\mathcal{M}_i$. The selection process ensures that communication efforts are focused on the most relevant and beneficial peers, optimizing model performance and enhancing training efficiency.

%Inspired by multi-task learning, we have adopted a partial-freeze training method in our subsequent model training sessions. This approach acknowledges that a model's core components, combined with different headers, can predict different tasks. We consider heterogeneity as a multi-tasking environment where different data distributions represent different tasks. Thus, we propose the partial-freeze method to accelerate and enhance training effectiveness. %Particularly, we divide the model parameters into two parts: feature extraction and the header. The feature extraction part is more generalized, encompassing the model's convolutional layers and the linear layers that do not make the final output decisions. These parameters are aggregated with those from other neighbors. The header represents each local client model's last multiple linear layers, which is entirely personalized. It does not participate in any form of aggregation and aligns with local data characteristics.%

After completing peer selection, the feature extraction layers are aggregated from each selected peer. The training then proceeds in two phases. In the first phase, the header layers are frozen, and the feature extraction layers are trained. This targeted training helps to enhance the model's ability to accurately interpret and process input data. Once this training phase is complete, the trained parameters can immediately be dispatched to peers that have already made requests. In the second phase, the feature extraction layers are frozen, and the training focuses on the header layers. This phase is dedicated to fine-tuning the header, which is responsible for making the final decisions and classifications based on the processed features. The overall training approach allows each component of the model to be optimized for its specific role, enhancing the overall performance and efficiency of the distributed learning system. The details of our proposed framework are shown in Algorithm \ref{alg:1}.

\begin{algorithm}[htb]
\caption{Data Fusion Algorithm}
\label{alg:1}
\SetAlgoLined
\KwIn{Total number of clients $M$; Local input data $D$; Total communication rounds $T$; Number of local training iterations $K_e$ and $K_h$; Communication cost $C$}
\KwOut{Personalized local trained models $e_i$ and $h_i$}

Initialize each local client's header parameters $h_{i,0}$, feature extraction parameters $e_{i,0}$, peer recency array $t$, loss array \(l\), communication cost score $c$ based on $C$ and peers information collector $n_{i}$

\For{$ t = 1$ \KwTo $T$}{
    \For{client $i$ in parallel \KwTo $M$}{

        Calculate the $\mathcal{S}_{i,j} = s_p(\alpha s_l - s_d + c)$ by Eq. (\ref{eq:sl}), (\ref{eq:sd}), and (\ref{eq:st})

        Construct the selected peers set $\mathcal{M}_i \in \{\mathcal{S}_{i,j} > s^*\}$

        Receive selected peer's parameters and get the aggregated feature extraction  parameters $e_i = \sum_{j \in \mathcal{M}_i}e_j$

        Update the loss array $l$
        
        \For{k = 1 \KwTo $K_e$}{
            Sample a batch of data $(x,y)$ from local dataset.

            Update feature extraction parameters $e_i$:
            $e_{i}^{t,k+1} = e_i^{t,k} - \eta_e \nabla_e  L(({h_i^t}^{f},e_i^t);x,y)$
        }\textbf{end for}

        Broadcast the updated $e_i$

        \For{k = 1 \KwTo $K_h$}{
            Sample a batch of data $(x,y)$ from local dataset.

            Update header parameters $e_i$:
            $h_{i}^{t,k+1} = h_{i}^{t,k} - \eta_h \nabla_h  L((h_i^t,{e_{i}^t}^{f});x,y)$
        }\textbf{end for}

        Update the peer recency array $t$
        
    }\textbf{end for}
}\textbf{end for}
\end{algorithm}

\section{Experiments}
In this section, we conduct extensive experiments to verify the effectiveness of the proposed PFedDST algorithm in scenarios characterized by data heterogeneity and computation resources heterogeneity. These experiments are designed to evaluate how well PFedDST handles diverse datasets and varying computational resources across different nodes in a distributed system.

\subsection{Experimental Setup}
To evaluate the performance of the proposed algorithm, we use the CIFAR-10 and CIFAR-100 datasets, which are real-life image classification datasets containing images distributed across 10 and 100 classes, respectively. These datasets are commonly used in machine learning research to benchmark image classification algorithms and are particularly useful for assessing performance in heterogeneous data distribution scenarios. The data for each dataset is partitioned in a Pathological distribution manner, intended to simulate a realistic scenario where each client may have access to only a limited subset of the total classes. Specifically, for CIFAR-10, we sample 2 classes from the total of 10 for each client. Similarly, for CIFAR-100, each client is assigned 5 classes from the total of 100. This partitioning method ensures that each client's training and testing data are distributed according to the same class subset, which introduces challenges typical of federated learning environments where data may not be identically and independently distributed across clients.

To ensure a fair comparison across all methods, we maintain consistent experimental conditions for each baseline. The experiments are conducted over 500 communication rounds involving 100 clients. Each client in the federated learning setup communicates with 10 neighbors, and similarly, in the PFedDST method, 10 clients are also chosen at each communication round. The client sampling ratio is set at 0.1. The training involves using a batch size of 128. For the PFedDST method, the feature extraction part is trained for 5 epochs per round, matching the training duration of other baselines. The header part is only trained for 1 epoch per round to reduce computational overhead. All methods employ Stochastic Gradient Descent (SGD) as the optimizer, with a learning rate 0.1. Additionally, all methods implement a decay rate of 0.005 and a local momentum of 0.9 to optimize the convergence and stability of training. The communication cost is equal between each client.

\subsection{Experimental Evaluation}

We assess our proposed methods against current state-of-the-art baselines in PFL. The evaluation includes centralized federated learning methods such as FedAvg \cite{mcmahan2017communication}, FedPer\cite{arivazhagan2019federated} and FedBABU \cite{oh2021fedbabu}, and decentralized federated learning methods such as DFedAvgM \cite{sun2022decentralized}, Dis-PFL\cite{dai2022dispfl}, and DFedPGP \cite{liu2024decentralized} and reproduced the result of \cite{liu2024decentralized}. Each method is tested using a ResNet-18 architecture. In our setup for partial PFL methods, the header layers are personalized for complex pattern recognition, while the remaining layers are shared for feature extraction. Our primary evaluation metric is personalized test accuracy, which aligns with our goal of addressing the challenges in PFL.

\begin{figure}
    \centering
    \includegraphics[width=1\linewidth]{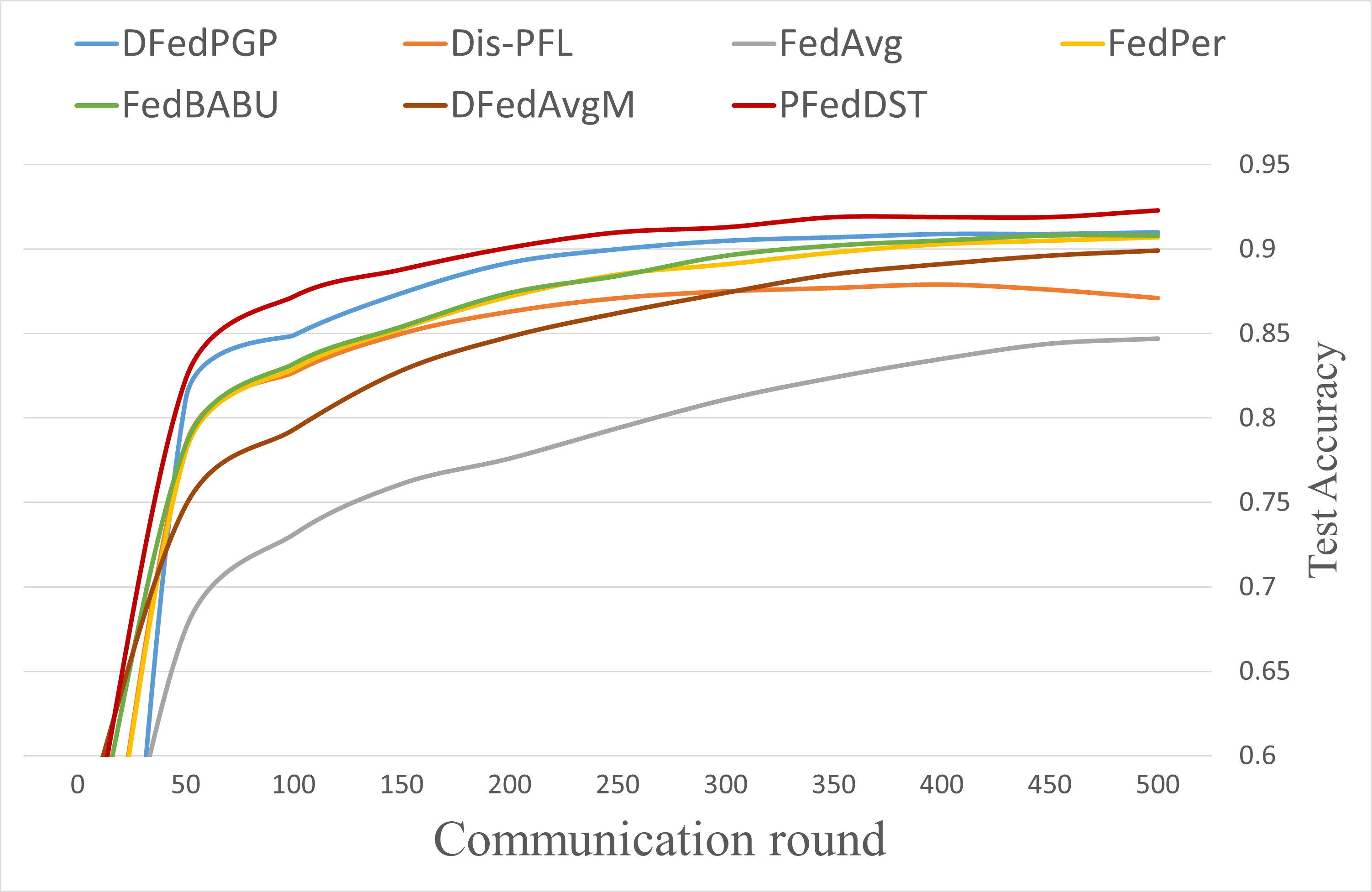}
    \caption{Test accuracy on CIFAR-10 }
    \label{fig:cifar10}
\end{figure}

\begin{figure}
    \centering
    \includegraphics[width=1\linewidth]{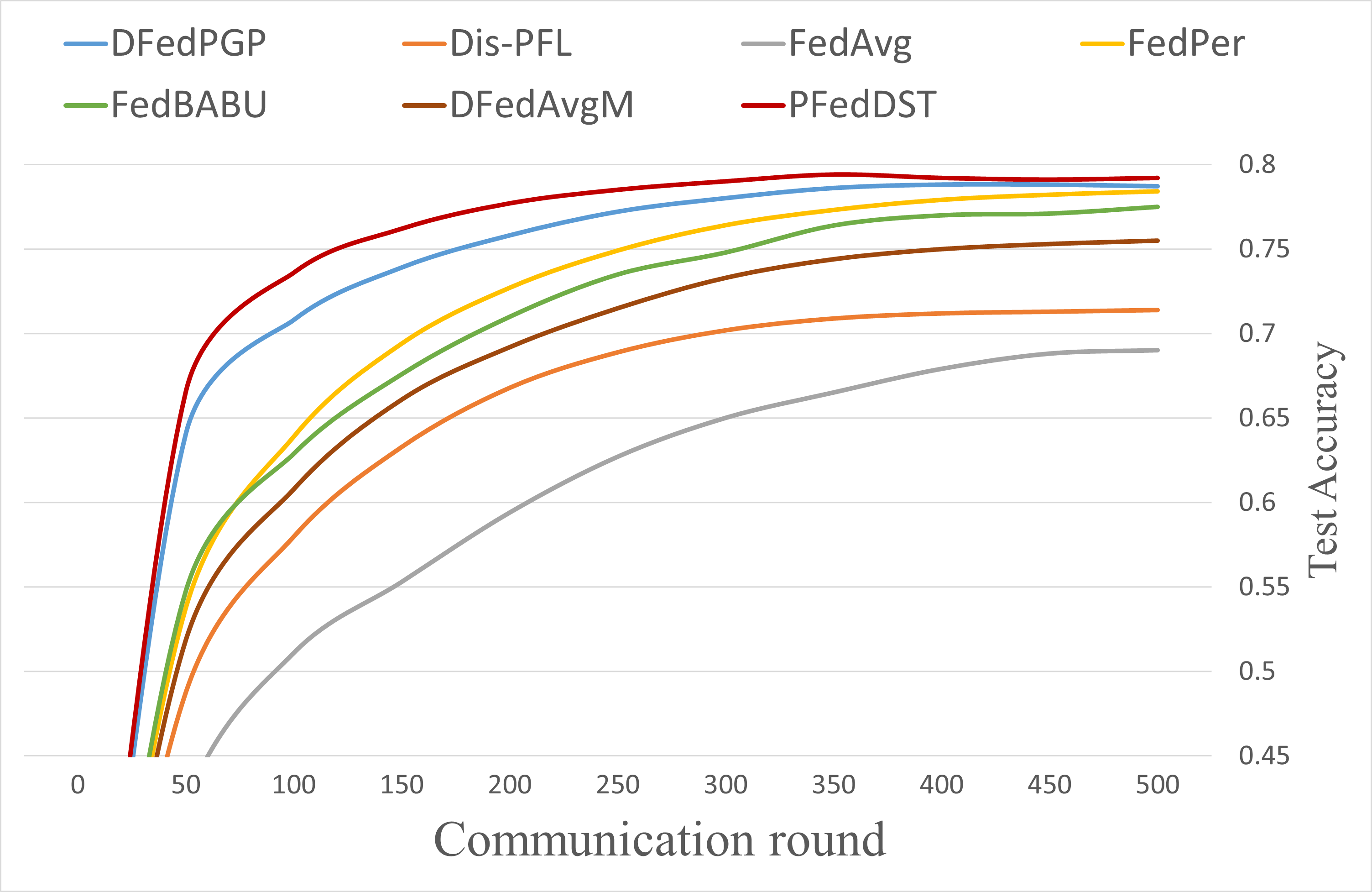}
    \caption{Test accuracy on CIFAR-100 }
    \label{fig:cifar100}
\end{figure}

As presented in Figure \ref{fig:cifar10} and \ref{fig:cifar100}, the proposed Personalized Federated Learning Decentralized Selection Training (PFedDST) shows \textit{superior stability and performance} over baseline methods across diverse datasets and scenarios of data heterogeneity. Specifically, on the CIFAR-10 dataset, PFedDST achieves a remarkable accuracy of \textbf{92.25\%}, outperforming the nearest baseline method, by \textbf{1.0\%}. On the CIFAR-100 dataset, DFedPGP leads with an accuracy of \textbf{79.41\%}, which is at least \textbf{0.7\%} higher than other baseline methods. The implementation of a communication protocol based on a directed graph allows clients to flexibly select their peers, thus facilitating the choice of pertinent information for their local training processes.

In Table 2, we present the learning curves illustrating the convergence speeds of the methods compared. PFedDST has the \textit{fastest convergence} among the methods tested, which benefited from the peers selection algorithm. Notably, DFedPGP demonstrates that a convergence rate is much better than other methods in both CIFAR-10 and CIFAR-100 scenarios. 

Compared to other methods, PFedDST further optimizes the clients' communication and aggregation. It enhances convergence speed and generalization capability by selecting peers based on their relevance scores, ensuring a more balanced choice of beneficial peers. Additionally, the use of a partially frozen training approach speeds up the training process and enhances transfer efficiency, which minimize the cost consumption while maximizing the information gain. 

\begin{table}[ht]
\centering
\caption{The required communication rounds when achieving the
target accuracy (\%).}
\label{exp_mnist_regular_cnn2}
\begin{tabular}{>{\centering\arraybackslash}m{2.3cm} 
                |>{\centering\arraybackslash}m{2.3cm} 
                |>{\centering\arraybackslash}m{2.3cm}}
\toprule
\textbf{Method} & \textbf{CIFAR-10 (target acc is 90)} & \textbf{CIFAR-100  (target acc is 75)} \\ 
\midrule
FedAvg\cite{mcmahan2017communication}          & - & -  \\
FedPer\cite{arivazhagan2019federated}          & 350 & 254  \\
%FedRep\cite{collins2021exploiting}          & 323 & 225 \\
FedBABU\cite{oh2021fedbabu}         & 321 & 306  \\
%Ditto\cite{li2021ditto}           & - & - \\
DFedAvgM\cite{sun2022decentralized}        & 462 & 399  \\
%OSGP\cite{assran2019stochastic}            & 442 & 342 \\
Dis-PFL\cite{dai2022dispfl}         & - & -  \\
DFedPGP\cite{liu2024decentralized}         & 238 & 178 \\
\textbf{PFedDST} & \textbf{184}  & \textbf{133} \\
\bottomrule
\end{tabular}
\end{table}
\section{Discussion}
 Our framework is designed to prioritize the selection of the most beneficial communication peers and utilize partial personalization, ensuring optimal performance and efficiency in distributed learning scenarios. The simulation demonstrates that our method outperforms the current state-of-the-art methods in terms of accuracy and convergence rate. This improvement is more distinct as the model complexity grows, data heterogeneity intensifies, and the number of clients increases.
 
 This enhancement is attributed to the combined training of a fully personalized header and a shared feature extraction layer, supplemented by an effective benefit selection strategy. Initially, we implement a partially frozen training method. During local optimization, the header is frozen while the feature extraction layers are actively trained. Upon completion, the trained component is shared with the required peers, and the previously frozen sections are then unfrozen for further training. This method diverges from traditional training approaches by reducing the number of model parameters trained and communicated, enabling faster training completion. Additionally, it promotes stable parameter optimization and minimizes gradient conflicts. Secondly, we employ a score selection strategy, evaluating potential communication partners across various dimensions, including loss, selection frequency, communication costs, and task similarity. This comprehensive scoring method facilitates the identification of the most suitable partners for exchange, consequently improving the overall training outcomes by increasing accuracy and speeding up convergence. A notable feature of PFedDST is robustness. This selection mechanism automatically filters out potential attackers and clients with noisy data by measuring header distances, improving the robustness of the local model aggregation.
\section{Conclusion}

In conclusion, we propose a unified decentralized federated learning selection framework PFedDST for personalization, fast convergence, privacy, robustness, and communication efficiency within distributed learning environments. By employing score selection score based on loss, peer recency, and task similarity on decentralized devices, we offer the PFedDST that enhances the ability to communicate with beneficial peer models while ensuring a fast convergence rate and privacy. Theoretical findings and experimental results show that our method achieved a faster convergence rate and higher model accuracy compared to other state-of-the-art methods.

% \begin{table}[ht]
% \caption{Comparisons between training methods on CNN model for MNIST dataset with non-IID data distribution.}
% \begin{tabular}{p{2.3cm}p{2.3cm}p{2.3cm}}
% \hline
% Method   & CIFAR-10 &  CIFAR100 \\ 
% \hline
%     Local & 85.16 $\pm$ .18 & 71.34 $\pm$ .46  \\
%     FedAvg & 85.04 $\pm$ .11 & 69.05 $\pm$ .43  \\
%     FedPer & 90.94 $\pm$ .24 & 78.48 $\pm$ .93  \\
%     FedRep & 91.09 $\pm$ .12 & 78.77 $\pm$ .19 \\
%     FedBABU & 91.28 $\pm$ .15 & 77.50 $\pm$ .33  \\
%     Ditto & 84.96 $\pm$ .40 & 69.48 $\pm$ .45 \\
%     DFedAvgM & 90.23 $\pm$ .97 & 75.89 $\pm$ .65  \\
%     OSGP & 90.72 $\pm$ .08 & 76.70 $\pm$ .59 \\
%     Dis-PFL & 88.19 $\pm$ .47 & 71.79 $\pm$ .42  \\
%     DFedPGP & 91.26 $\pm$ .05 & 78.78 $\pm$ .41 \\
%     \textbf{PFedDSST} & \textbf{92.25 $\pm$ .74}  & \textbf{79.41 $\pm$ .57} \\
%     \hline
% \end{tabular}
% \label{exp_mnist_regular_cnn}
% \end{table}

\bibliographystyle{IEEEtran}
\bibliography{refer.bib}

\end{document}